\documentclass[conference]{IEEEtran}
\IEEEoverridecommandlockouts
\usepackage{cite}
\usepackage{amsmath,amssymb,amsfonts}
\usepackage{algorithmic}
\usepackage{graphicx}
\usepackage{textcomp}
\usepackage{xcolor}
\usepackage{enumitem}
\usepackage{subcaption}
\usepackage{booktabs}
\usepackage{tabularx}
\usepackage{multirow}
\usepackage{cuted}
\usepackage{xurl}
\usepackage{hyperref}
\usepackage[font=footnotesize]{caption}

\IEEEaftertitletext{\vspace{-2\baselineskip}}
\begin{document}

\newcommand{\JS}[1]{{\color{red}[JS] #1}}
\newcommand{\D}{\mathbf{D}}
\newcommand{\SO}{SO(3)}
\newcommand{\vect}{\text{vec}}
\newcommand{\SE}{SE(3)}

\newcommand{\fg}{f_{G}}
\newcommand{\eg}{e_{_G}}
\newcommand{\ev}{e_V}
\newcommand{\Ad}{\text{Ad}}
\newcommand{\tr}{\text{tr}}
\newcommand{\se}{\mathfrak{se}(3)}
\newcommand{\so}{\mathfrak{so}(3)}
\newcommand{\ad}{\text{ad}}

\newcommand{\IP}[2]{\langle {#1}, {#2} \rangle}

\title{\LARGE \bf 
Precise $\SE$ End-Effector Tracking in \\ Whole-Body Humanoid Control
}

\author{
Joohwan Seo\textsuperscript{1,2}, Xiaofeng Guo\textsuperscript{2}, Jinkun Cao\textsuperscript{2},\\ Roberto Horowitz\textsuperscript{1}, Rocky Duan\textsuperscript{2}, Guanya Shi\textsuperscript{2,3,$\dagger$}, Koushil Sreenath\textsuperscript{1,2,$\dagger$}
\thanks{$^{1}$University of California, Berkeley, $^{2}$Amazon FAR, $^{3}$Carnegie Mellon University}
\thanks{$^\dagger$Equal Advising}
}

\maketitle

\begin{strip}
    \centering
    \includegraphics[width=0.98\textwidth]{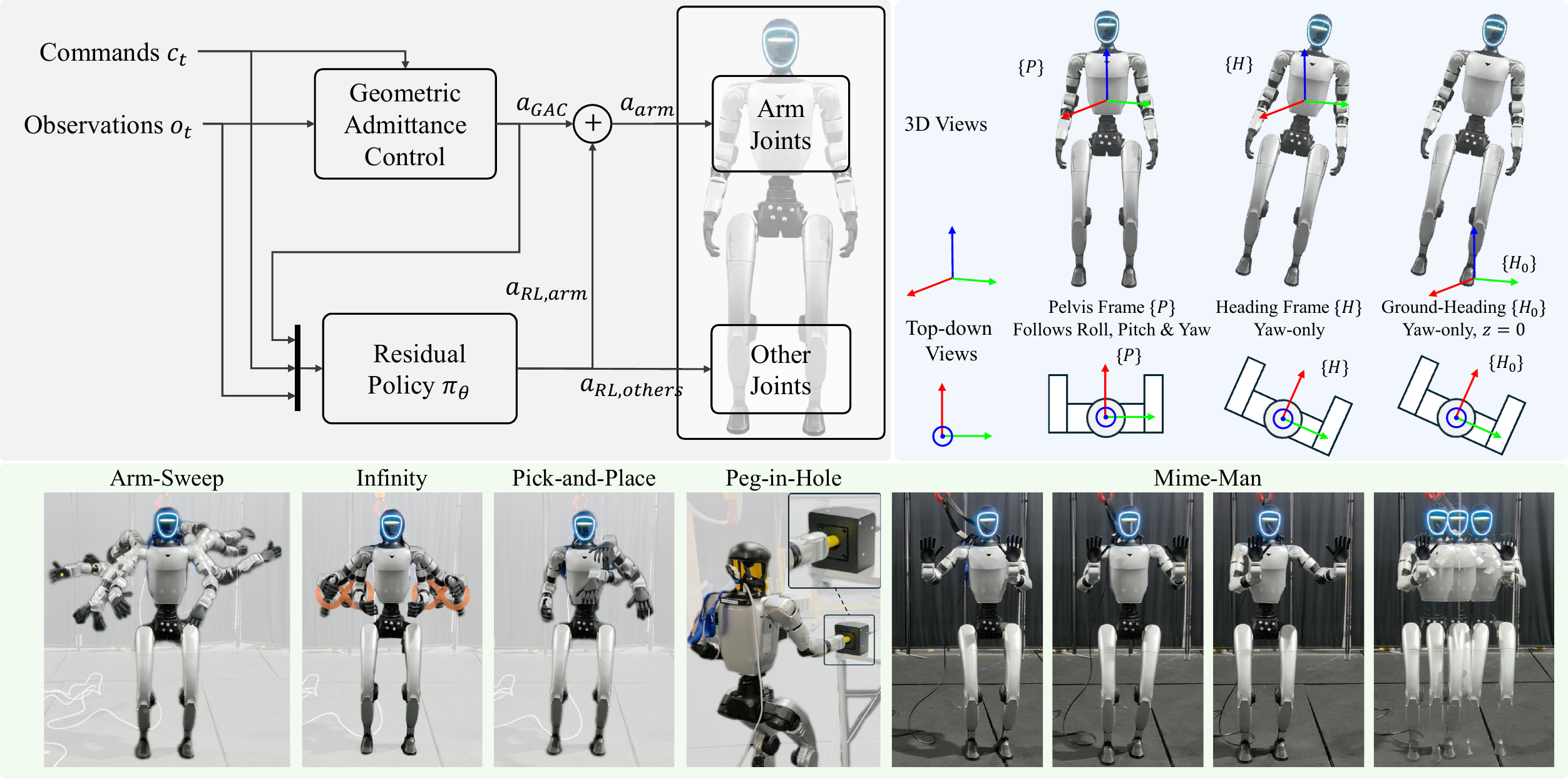}
    \vspace{-0.3em}
    \captionof{figure}{\footnotesize 
    Overview of the proposed framework for precise $\SE$ end-effector tracking in whole-body humanoid control.
    (a) Our framework, \textbf{ResGAC}, consists of a geometric admittance controller (GAC) that provides the nominal end-effector command and a residual RL policy that learns arm corrections while controlling the remaining whole-body joints.
    (b) We consider the pelvis $P$, heading $H$, and ground-attached heading $H_0$ frames. The command defined in $H$ removes pelvis roll and pitch, while $H_0$ additionally removes pelvis heave by projecting the heading frame onto the ground, isolating planar $(x,y,\psi)$ motion from the manipulation task.
    (c) The framework is evaluated on diverse $\SE$ end-effector pose tracking and manipulation tasks, including arm-sweep, infinity, pick-and-place, peg-in-hole, and a Mime-man demo for world-frame end-effector tracking. 
    }
    \label{fig:ResGAC_main}
\end{strip}
\vspace{-15pt}

\begin{abstract}
Precise end-effector tracking during humanoid whole-body motion is challenging due to floating-base oscillations, gravity, dynamic coupling, and locomotion-induced disturbances. We propose \textit{ResGAC}, a whole-body humanoid controller for precise end-effector pose tracking that combines geometric admittance control (GAC) with residual reinforcement learning. GAC provides structured $\SE$ task-space feedback and generates nominal arm joint-position targets, while residual RL compensates for unmodeled dynamics and coordinates locomotion and balance in the shared joint-position action space. The left-invariant geometric formulation allows the same GAC law to be used across manipulation reference frames. This enables the use of a ground-attached heading frame that preserves planar locomotion while removing pelvis roll, pitch, and heave from the manipulation reference, thereby reducing reference-induced end-effector motion during locomotion. 
ResGAC is validated on a real Unitree G1 humanoid. Across four standing end-effector tracking benchmarks, ResGAC consistently outperforms representative baselines, including SONIC, achieving lower translational and rotational errors. Real-world experiments further demonstrate reduced propagation of pelvis motion to the desired end-effector pose using the proposed ground-attached heading frame. ResGAC achieves $90\%$ success in a standing peg-in-hole task compared with $50\%$ for SONIC, and accurate world-frame $\SE$ end-effector pose tracking during lower-body motion. Experimental videos are included in the supplementary material and are also available on the project website: \textcolor{blue}{\url{https://resgac.github.io/ResGAC-website/}}.
\end{abstract}


\section{Introduction}
Recent advances in learning-based humanoid control have enabled increasingly versatile whole-body behaviors through large-scale motion tracking \cite{tessler2024maskedmimic,liao2025beyondmimic,luo2026sonic, he2024learning, he2024omnih2o}. However, manipulation via humanoid robots requires more than motion naturalness: the robot needs to accurately track commanded end-effector poses while performing locomotion and maintaining balance. This is challenging because manipulation is coupled with base oscillations, gravity, whole-body dynamics, and foot-impact disturbances, particularly for end-effector tracking in $\SE$ where both translational and rotational precision are required.

Most humanoid controllers are built around dense motion-tracking objectives \cite{he2024learning,he2024omnih2o}, whereas manipulation tasks are often more naturally specified through sparse workspace commands. Recent works, such as \cite{yang2026handoff,cao2026hiwet,luo2026sonic}, therefore move toward task-space end-effector interfaces, allowing the controller to resolve the remaining whole-body motion. Related works further investigate direct end-effector regulation \cite{ben2025homie}, including $\SE$ pose tracking for mobile manipulation in quadrupeds \cite{portela2025whole} and end-effector stabilization against locomotion-induced disturbances in humanoids \cite{li2025hold,jang2025seec}. Recent approaches extend this task-space perspective to physical interaction through controllable compliance and force-adaptive control \cite{chen2025chip,zhang2025falcon}. Nevertheless, high-precision tracking of the end-effector pose in $\SE$, jointly accounting for translational and rotational accuracy during humanoid whole-body motion, remains comparatively under-emphasized.

Residual learning provides a practical mechanism for combining structured priors with the robustness of learned whole-body control. Existing approaches employ different forms of such priors: A pretrained general motion-tracking policy is refined through residual RL in \cite{zhao2025resmimic}, while joint-space residual corrections around an inverse kinematics (IK)-derived kinematic prior are learned in \cite{cao2026hiwet}. An IK-derived joint-position reference is similarly corrected in joint space, while model-based feedforward and gravity-compensation torques are separately incorporated into the final control command \cite{cheng2025rambo} in quadruped loco-manipulation. 
In \cite{jang2025seec}, operational-space compensation and task-control torques are analytically derived but used only to guide the training of a policy whose deployed action is represented in joint space.

The choice of manipulation reference frame is another important consideration for whole-body end-effector control. A natural choice is the pelvis frame. However, command representation on the pelvis frame transmits periodic pelvis roll, pitch, and vertical motion during locomotion to the desired end-effector pose. Recent humanoid-control implementations \cite{zakka2026mjlab,luo2026sonic} address part of this issue using a heading frame, which retains the floating base position and yaw while removing its roll and pitch.

Motivated by these observations, we propose ResGAC, a framework for precise $\SE$ end-effector tracking 
in whole-body humanoid control. An explicit geometric admittance controller (GAC) \cite{seo2023geometric,seo2023contact} is formulated on the floating base to generate nominal joint-position targets from $\SE$ task-space feedback, and residual RL is applied in the same joint-position action space, allowing the learned correction to be directly composed with the structured feedback command while compensating for unmodeled dynamics and coordinating locomotion and balance. In addition, the left-invariant geometric formulation \cite{seo2023contact} allows the same GAC law to be utilized across manipulation reference frames. We introduce a ground-attached heading frame that additionally prevents vertical pelvis motion from propagating to the desired end-effector pose through the manipulation reference while retaining the planar motion of the humanoid.

Our main contributions are as follows:
%
\begin{itemize}[leftmargin=*]
    \item We propose \textbf{ResGAC}, integrating floating-base geometric admittance control (GAC) with residual RL in a shared joint-position action space, combining geometric $\SE$ task-space feedback with learned corrections for whole-body control.
    \item The left-invariant GAC formulation is extended across manipulation reference frames, and a ground-attached heading frame is introduced to 
    prevent pelvis roll, pitch, and heave from propagating to the desired end-effector pose while preserving planar locomotion.
    \item ResGAC is validated on the real-world Unitree G1 through end-effector position-and-orientation tracking, static arm-holding during walking, peg-in-hole manipulation, and world-frame tracking, consistently outperforming representative baselines, including SONIC \cite{luo2026sonic}.
\end{itemize}

\section{Background and Problem Formulation}
This section introduces the basic notation for poses, twists, and Jacobian matrices used throughout the paper and formulates the whole-body end-effector tracking problem.

\subsection{Lie Group and Frame Notation} \label{sec:notation}
We consider three types of frames for floating-base humanoid end-effector control: the world frame $\{W\}$, a manipulation reference frame $\{R\}$ in which the end-effector command and controller are defined, and the end-effector frame $\{E\}$. The manipulation reference
frame is chosen from the pelvis frame $\{P\}$, heading frame $\{H\}$,
or ground-attached heading frame $\{H_0\}$ as illustrated in Fig.~\ref{fig:ResGAC_main}(b). The world frame is chosen to have fixed position and orientation. Hereafter, we omit braces around frame symbols unless needed for clarity.

The end-effector pose $E$ with respect to the manipulation reference frame $R$ is represented by the homogeneous matrix representation
\begin{equation}\label{eq:homo}
    g_{RE}
    =
    \begin{bmatrix}
        R_{RE} & p_{RE} \\
        0 & 1
    \end{bmatrix}
    \in \SE,
\end{equation}
where $R_{RE}\in\SO$ and $p_{RE}\in\mathbb{R}^3$. At each control step, $g_{RE}$ is obtained directly from forward kinematics with respect to the selected manipulation reference frame.
For notational compactness, $E$ is used generically for both the left
and right end effectors, such that $g_{RE}$ represents the pair
$(g_{RE_l},g_{RE_r})$.

The relative end-effector velocity expressed in the end-effector frame is represented by the body twist
\begin{equation}
    V_{RE}^{E}
    =
    \begin{bmatrix}
        v_{RE}^{E} \\
        \omega_{RE}^{E}
    \end{bmatrix}
    \in \mathbb{R}^{6},
\end{equation}
where $v_{RE}^{E}$ and $\omega_{RE}^{E}$ denote the translational and angular velocities of $E$ with respect to $R$ expressed in $E$, respectively. We also use hat- and vee-maps to map a vector in $\mathbb{R}^3$ to a skew symmetric matrix in $\so$, a Lie algebra of $SO(3)$, and vice-versa, i.e.,
\begin{equation}
    (\cdot)^\wedge: \mathbb{R}^3 \mapsto \so, \quad (\cdot)^\vee: \so \mapsto \mathbb{R}^3.
\end{equation}
The Adjoint map transforms twist coordinates between frames. 
For $g  = (p,R) \in \SE$, its matrix representation is given as
\begin{equation}
    \Ad_g = 
    \begin{bmatrix}
        R & \hat{p}R \\ 0 & R
    \end{bmatrix} \in \mathbb{R}^{6\times6}.
\end{equation}
The corresponding body-frame Jacobian $J_{RE}^{E}$ is defined through
\begin{equation}\label{eq:body_jacobian}
    V_{RE}^{E}
    =
    J_{RE}^{E}\dot{q},
\end{equation}
where $\dot{q}$ denotes the generalized velocity associated with the associated kinematic chain. 
For constructing the relative quantities from the kinematic interfaces, we additionally define the world-aligned Cartesian frames $\{C_R\}$ and $\{C_E\}$, whose positions coincide with $R$ and $E$, respectively, while their orientations are aligned with $W$. Accordingly, $V_{WR}^{C_R}$ and $V_{WE}^{C_E}$ denote the world-relative velocities of the reference and end-effector frames expressed in their corresponding world-aligned Cartesian frames. For further mathematical background on Lie-group notation, readers are referred to \cite{murray1994mathematical}.

\subsection{Problem Formulation} \label{sec:problem_formulation}
Whole-body humanoid control addressed in this paper involves simultaneous locomotion and dual-arm end-effector pose tracking. The command at time $t$ is defined as
\begin{equation}\label{eq:command}
    c_t =
    \left(
        v_{d,x}^{P},
        v_{d,y}^{P},
        \omega_{d,z}^{P},
        z_d^{P},
        g_{RD}
    \right),
\end{equation}
where $v_{d,x}^{P}$, $v_{d,y}^{P}$, and $\omega_{d,z}^{P}$ denote the desired planar locomotion velocities, $z_d^{P}$ the desired pelvis height, and $g_{RD}$ denotes the desired left and right end-effector poses represented in the manipulation reference frame $R$, where $\{D\}$ stands for the desired frame. Similar to the definition of the end-effector, $g_{RD}$ compactly represents the pair $(g_{RD_l},g_{RD_r})$.
The objective is to accurately track the commanded end-effector trajectories in both position and orientation while executing locomotion and maintaining balance.

The humanoid is controlled through joint-position targets tracked by low-level PD controllers, whereas the manipulation objective is naturally defined in task space on $\SE$. The nominal task-space feedback is therefore mapped to a joint-position target, such that the residual correction is applied at the same low-level control interface. The nominal controller is recovered when the residual is zero, while the learned action provides a bounded correction to the nominal joint-position command.

The manipulation reference frame also directly affects the desired world-frame trajectory:
\begin{equation}\label{eq:world_target}
    g_{WD}=g_{WR}\cdot g_{RD}.
\end{equation}
Thus, motion of reference frame $R$ is propagated to the end-effector target even when $g_{RD}$ is fixed. In particular, a pelvis-frame representation transmits locomotion-induced roll, pitch, and vertical oscillations to the desired end-effector motion. A suitable manipulation reference should therefore retain task-relevant planar base motion while preventing pelvis motion from propagating to the desired end-effector pose.

\section{Method}
This section presents the ResGAC framework. 
We first formulate geometric admittance control (GAC) for a floating-base humanoid and combine its nominal arm command with residual RL. We then introduce the manipulation reference-frame formulation and describe the training procedure used to learn the whole-body policy.

At each control step, GAC generates nominal arm joint-position targets, while the residual policy corrects the arm targets and controls the remaining whole-body joints, as illustrated in Fig.~\ref{fig:ResGAC_main}(a).
\subsection{Geometric Admittance Control on the Floating Base} \label{sec:GAC}
The floating-base GAC formulation is developed in two steps. We first derive the relative Jacobian and velocity with respect to the manipulation reference frame and then formulate the GAC joint-position command with null-space handling.
\subsubsection{Calculation of Jacobian Matrix}
The body-frame relative Jacobian $J_{RE}^{E}$ and relative end-effector velocity $V_{RE}^{E}$ are required for GAC \cite{seo2023contact}. For a floating-base humanoid, these quantities describe the motion of the end-effector relative to a moving reference frame $R$, whereas the kinematic interfaces used in training (\texttt{IsaacSim}) and deployment (\texttt{pinocchio} \cite{carpentier-sii19}) primarily provide Jacobians and velocities of individual frames relative to the world or do not provide the relative quantities between different frames. We therefore first construct the relative quantity in the world-aligned Cartesian-frame representation and subsequently transform it into the end-effector body frame.

The relative end-effector velocity expressed in $C_E$ is given by
\begin{equation} \label{eq:velocity}
    V_{RE}^{C_E} = V_{WE}^{C_E} - V_{WR}^{C_E}.
\end{equation}
Since $V_{WR}^{C_R}$ is directly available from kinematic interface, it is transformed to $C_E$ using $g_{C_E C_R} = g_{WC_E}^{-1} ~ g_{WC_R}$:
\begin{equation}
    V_{WR}^{C_E} = \Ad_{g_{C_E C_R}} V_{WR}^{C_R} = \begin{bmatrix}
        I & -({p_{WC_E} \!-\! p_{WC_R}})^\wedge \\
        0 & I
    \end{bmatrix} V_{WR}^{C_R}.
\end{equation}
Accordingly, the relative Jacobian is obtained as
\begin{align} \label{eq:ref_jacobian}
    J_{RE}^{C_E} = J_{WE}^{C_E} - J_{WR}^{C_E} = J_{WE}^{C_E} - \Ad_{g_{C_E C_R}}J_{WR}^{C_R}.
\end{align}
The body-frame Jacobian is then given by
\begin{equation}
    J_{RE}^E = \Ad_{g_{EC_E}} J_{RE}^{C_E} = \begin{bmatrix}
        R_{WE}^\top & 0 \\
        0 & R_{WE}^\top
    \end{bmatrix} J_{RE}^{C_E},
\end{equation}
from which the relative body-frame velocity is calculated as
\begin{equation} \label{eq:eef_velocity}
    V_{RE}^E = J_{RE}^E \dot{q}.
\end{equation}
The relative Jacobian in \eqref{eq:ref_jacobian} contains all joints contributing to the end-effector motion relative to ${R}$, including the torso joints. 
While the full Jacobian is used to evaluate $V_{RE}^{E}$, GAC is applied only through the arm-joint columns, leaving the remaining joints to be fully handled by the RL policy. 

\subsubsection{Geometric Admittance Control with Null-Space Handling}
Geometric Admittance Control (GAC) \cite{seo2023contact} is derived from the Geometric Impedance Control formulation \cite{seo2023geometric}. 
Unlike its original use for compliant control, GAC is used here as a geometric task-space tracking controller that generates joint-position targets from the $\SE$ pose error without measured external wrench. The resulting GAC command can be interpreted as a differential IK step. 
In addition to the resulting joint target, ResGAC provides the pose tracking error and relative end-effector velocity to the residual policy, as described in Sec.~\ref{sec:residual_rl}.
Under the second-order approximate GAC closed-loop dynamics,
these quantities form an approximately Markovian representation of the task-space tracking state.
For compactness, $V_{RE}^E = V$ and the arm-restricted $J_{RE}^E = J$ are used below. The desired task-space dynamics are
\begin{equation}
    M\dot{V} + K_d V + \fg(g_{RE},~g_{RD}) = 0,
\end{equation}
where $M,K_d \in \mathbb{R}^{6\times6}$ are symmetric positive definite matrices, and $\fg(\cdot,\cdot)$ denotes the elastic wrench, i.e., P-control action, on $\SE$ \cite{seo2023geometric}:
\begin{equation} \label{eq:desired_dynamics_GAC}
    \fg(g,g_D) = \begin{bmatrix}
    f_p \\ f_R
    \end{bmatrix} =\begin{bmatrix}
    R^\top R_D K_p R_D^\top (p - p_D)\\
    (K_R R_D^\top R - R^\top R_D K_R)^\vee
    \end{bmatrix},
\end{equation}
where $g=(p,R)$ and $g_D=(p_D,R_D)$ compactly denotes $g_{RE}$ and $g_{RD}$, respectively. Moreover, $K_p,K_R\in\mathbb{R}^{3\times3}$ are positive-definite translational and rotational stiffness matrices, respectively. The geometric feedback is left-invariant: for any $g_l\in\SE$,
$\fg(g_lg,g_lg_D)=\fg(g,g_D)$.
Hence, a common change of reference coordinate frame leaves the feedback law unchanged.

The desired dynamics are discretized to obtain the desired body-frame velocity $V_D$:
\begin{equation}
    V_{D,t} = V_t + \Delta t \cdot M^{-1}\left( - f_{G,t} - K_d V_t \right),
\end{equation}
where $\Delta t$ is a sampling time.
The desired velocity is subsequently mapped to joint space using a damped pseudo-inverse, with null-space regulation to handle redundancy of the 7-DoF manipulator.
\begin{equation}
    \begin{split}
        \dot{q}_{D,t} &= J_{\lambda,t}^\dagger V_{D,t} + \dot{q}_{\mathrm{null},t} \\
        q_{D,t} &= q_{\mathrm{arm},t} + \Delta t \cdot \dot{q}_{D,t},
    \end{split}
\end{equation}
where $q_{\mathrm{arm},t}$ and $q_{D,t}$ denote the current and desired arm joint positions, respectively. In addition,
\begin{equation}
    \begin{split}
        J_\lambda^\dagger &= J^\top ( J J^\top + \lambda I)^{-1} \\
        \dot{q}_{\mathrm{null},t} &= (I - J_\lambda^\dagger J) \cdot (-k_{\mathrm{null}}(q_t - q_{\mathrm{rest}})),
    \end{split}
\end{equation}
and $q_{\mathrm{rest}}$ and $k_{\mathrm{null}}$ denote the nominal arm configuration and null-space gain, respectively. Finally, the resulting joint-position target is defined as the nominal GAC action,
\begin{equation}
    a_{\mathrm{GAC},t}=q_{D,t},
\end{equation}
which is directly combined with the residual policy introduced subsequently.

\subsection{Residual Reinforcement Learning} \label{sec:residual_rl}
ResGAC employs residual RL on top of the nominal GAC command. The GAC provides structured geometric feedback for the arms, while the RL policy learns bounded arm corrections and controls the remaining whole-body joints for locomotion and balance. In this way, dynamic effects not captured by the nominal controller, including gravity and whole-body coupling, can be compensated through learning.

The policy operates in the joint-position action space. Its output is partitioned into the arm and remaining joints as
\begin{equation}
a_{\mathrm{RL},t}
=
\pi_{\theta}(o_t)
=
[
a_{\mathrm{RL,arm},t}^\top,\;
a_{\mathrm{RL,other},t}^\top
]^\top.
\end{equation}
The applied arm action is
\begin{equation}
a_{\mathrm{arm},t} = a_{\mathrm{GAC},t} + s_r \cdot \operatorname{clip} \left(
a_{\mathrm{RL,arm},t}, -c_r, c_r \right),
\end{equation}
where $s_r$ and $c_r$ denote the residual scaling and clipping factors, respectively, 
while $a_{\mathrm{other},t}=a_{\mathrm{RL,other},t}$ is applied directly to the remaining joints.
Thus, zero residual action recovers the nominal GAC command, while the learned arm action remains a correction in the same joint-position action space. 

The nominal action $a_{\mathrm{GAC},t}$ is explicitly included in the policy observation, providing the residual policy with the structured command being corrected. The actor observation is
\begin{equation}
\begin{aligned}
o_t = \bigl(&
c_t, q_t, \dot{q}_t, g_t^{\mathrm{proj}}, \omega_t,
a_{\mathrm{GAC},t}, g_{DE,t}, \\
& p_{DE,t}^{\mathrm{sat}}, V_{RE,t}^{E},
a_{\mathrm{RL},t-1}, a_{\mathrm{RL},t-2}, \phi_t
\bigr).
\end{aligned}
\end{equation}
Here, $c_t$ is the task-space command defined in \eqref{eq:command},
$g_t^{\mathrm{proj}}$ and $\omega_t$ denote projected gravity and base angular velocity,
and $g_{DE,t}$ and $V_{RE,t}^{E}$ provide the end-effector pose error and relative body-frame velocity.
The pose error $g_{DE,t}$ and the pose command $g_{RD,t}$ \eqref{eq:command} are each represented by a $3$D translation and the $6$D rotation representation of \cite{zhou2019continuity}.
To retain sensitivity to small translational errors while limiting its range, an additional position feature is defined by
\begin{equation}
    p_{DE,t}^{\mathrm{sat}}
    =
    \tanh(100 \cdot p_{DE,t}).
\end{equation}

The gait state $\phi_t$ contains the sinusoidal phase variables used for locomotion \cite{Amazon_FAR_and_Abbeel_Holosoma, siekmann2021sim} and the two previous residual actions are included to provide the short-term action history required for the acceleration penalty.

An asymmetric actor-critic structure is used during training, where the critic is additionally provided with privileged simulation quantities, including pelvis linear velocity. This privileged information is available only to the critic and is not provided to the actor during training or deployment.

\subsection{Selection of Manipulation Reference Frame} \label{sec:reference_frame}

As discussed in Sec.~\ref{sec:problem_formulation}, the manipulation reference frame determines which components of floating-base motion are propagated to the end-effector command. We consider the pelvis frame $P$, heading frame $H$, and ground-attached heading frame $H_0$, illustrated in Fig.~\ref{fig:ResGAC_main}(b):
\begin{equation}
\begin{aligned}
p_{WH} &= p_{WP},            & R_{WH} &= R_z(\psi),\\
p_{WH_0} &= P_{xy}\cdot p_{WP},    & R_{WH_0} &= R_z(\psi),
\end{aligned}
\end{equation}
where $\psi$ is the pelvis heading angle, $P_{xy}=\operatorname{diag}(1,1,0)$, and $R_z(\psi)\in \SO$ denotes the rotation about the world $z$-axis associated with the heading angle $\psi$. 
Thus, $H$ removes pelvis roll and pitch, while $H_0$ additionally removes heave, preserving only the planar $(x,y,\psi)$ motion. Consequently, these excluded components of pelvis motion are not propagated to the desired world-frame end-effector pose through the reference transformation in~\eqref{eq:world_target}. Physical disturbances transmitted through whole-body dynamics, such as locomotion impacts, must still be compensated by the learned controller.


Let $f=R_{WP}e_x=[f_x,f_y,f_z]^\top$ denote the pelvis forward direction, where
$e_x=[1,0,0]^\top$, and also let Jacobian $J$ be partitioned as translational and rotational components $J_v$ and $J_\omega$. By differentiating
$\psi=\operatorname{tan}^{-1}(f_y/f_x)$, the heading Jacobian can be
shown to be
\begin{equation}
J_\psi
=
\begin{bmatrix}
-\frac{f_xf_z}{f_x^2+f_y^2} &
-\frac{f_yf_z}{f_x^2+f_y^2} &
1
\end{bmatrix}
J_{\omega,WP}^{C_P}.
\end{equation}
It then follows that the Jacobians for $H$ and $H_0$ are
\begin{equation}
J_{WH}^{C_H}
=
\begin{bmatrix}
J_{v,WP}^{C_P}\\
e_zJ_\psi
\end{bmatrix},
\qquad
J_{WH_0}^{C_{H_0}}
=
\begin{bmatrix}
P_{xy}J_{v,WP}^{C_P}\\
e_zJ_\psi
\end{bmatrix},
\end{equation}
where $e_z=[0,0,1]^\top$. These Jacobians replace
$J_{WR}^{C_R}$ in \eqref{eq:ref_jacobian}. Since pelvis angular
velocity is directly available from the IMU while linear velocity is not reliably measured at deployment, only the measured floating-base angular-velocity components are used in the actor-side reference-frame velocity calculation.

\vspace{-7pt}

\subsection{Training of ResGAC}
\subsubsection{Reward Function}
\begin{table}[t]
    \setlength\doublerulesep{0.5pt}
    \scriptsize
    \centering
    \setlength{\tabcolsep}{2.2pt}
    \renewcommand{\arraystretch}{1.00}
    \caption{
    Reward terms used for ResGAC training in the final curriculum phase.
    The total reward is $r_t=\sum_i w_i r_{i,t}$.
    $\exp(\cdot)$ denotes an exponential kernel. 
    Values correspond to the final training stage; curriculum-dependent
    changes are summarized separately.
    }
    \label{table:reward}
    \vspace{-4pt}

    \begin{tabularx}{\columnwidth}{
        >{\raggedright\arraybackslash\hsize=0.45\hsize}X
        >{\raggedright\arraybackslash\hsize=1.32\hsize}X
        >{\centering\arraybackslash\hsize=0.80\hsize}X
        >{\centering\arraybackslash\hsize=0.62\hsize}X
        >{\centering\arraybackslash\hsize=0.58\hsize}X
    }
    \toprule[1pt]\midrule[0.3pt]
    Group & Term & Form & $\sigma$ / Param. & $w_i$ \\
    \midrule\midrule

    \multirow{2}{*}{Manip.}
    & EEF position
        & $\exp{(L_2^2+L_1)}$ & $0.001/0.01$ & $4.0$ \\
    & EEF orientation
        &  $\exp{(SO(3))}$ & $0.02$ & $3.0$ \\

    \midrule

    \multirow{6}{*}{Loco.}
    & Linear velocity
        & $\exp{(L_2^2)}$ & $0.25$ & $2.0$ \\
    & Angular velocity
        & $\exp{(L_2^2)}$ & $0.25$ & $1.5$ \\
    & Swing-foot trajectory
        & $\exp{(L_2^2)}$ & $0.008$ & $5.0$ \\
    & Pelvis orientation
        & $\exp{(L_2^2)}$ & $0.04$ & $1.0$ \\
    & Torso orientation
        & $\exp{(L_2^2)}$ & $0.04$ & $1.0$ \\
    & Base height
        & $\exp{(L_2^2)}$ & $3.0$ & $1.0$ \\

    \midrule

    \multirow{17}{*}{Reg.}
    & Arm residual
        & $L_2^2$ & -- & $-0.05$ \\
    & EEF body twist
        & $L_2^2$ & -- & $-0.05$ \\
    & Arm action rate
        & $\|\Delta a\|^2$ & -- & $-0.5$ \\
    & Non-arm action rate
        & $\|\Delta a\|^2$ & -- & $-1.0$ \\
    & Mean-action accel.
        & $\|\Delta^2\mu\|^2$ & -- & $-0.10$ \\
    & Arm joint velocity
        & $L_2^2$ & -- & $-0.05$ \\
    & Pose deviation
        & Weighted $L_2^2$ & -- & $-1.0$ \\
    & Joint-position limit
        & Linear excess & $95\%$ & $-1.0$ \\
    & Foot proximity
        & Threshold & $0.15$ m & $-10$ \\
    & Foot orientation
        & $L_2$ & -- & $-5$ \\
    & Turn-in-place drift
        & Gated $L_2^2$ & -- & $-2.5$ \\
    & Heading drift
        & Gait-avg. $L_2^2$ & -- & $-10$ \\
    & Lateral drift
        & Gait-avg. $L_2^2$ & -- & $-2$ \\
    & Standing yaw rate
        & Gated $L_2^2$ & -- & $-2.5$ \\
    & Foot contact force
        & Threshold $L_2^2$ & $400$ N & $-2$ \\
    & Pre-touchdown velocity
        & Gated $L_2^2$ & $0.15$ m/s & $-2$ \\
    & Alive / termination
        & Indicator & -- & $10/-200$ \\
    \midrule[0.3pt]\bottomrule[1pt]
    \end{tabularx}
    \vspace{-4pt}
\end{table}
The reward jointly optimizes end-effector tracking and locomotion while regularizing undesirable whole-body behavior:
\begin{equation}
    r_t
    =
    r_t^{\mathrm{manip}}
    +
    r_t^{\mathrm{loco}}
    +
    r_t^{\mathrm{reg}}
\end{equation}
where the complete reward terms and their final-stage weights are summarized in Table~\ref{table:reward}. For precise manipulation tracking, let $g_{DE}=(p_{DE},R_{DE})$ denote the end-effector pose error. The position reward combines squared-$L_2$ and $L_1$ errors as
\begin{align}
    &r_{\mathrm{pos},t}
    =
    \exp\left(
    -\frac{e_{2,t}}{\sigma_2}
    -\frac{e_{1,t}}{\sqrt{3}\sigma_1}
    \right), \; \text{where} \\
    &e_{2,t}
    =
    \operatorname{mean}_{i\in\{l,r\}}
    \|p_{DE_i,t}\|_2^2,
    \;\;
    e_{1,t}
    =
    \operatorname{mean}_{i\in\{l,r\}}
    \|p_{DE_i,t}\|_1. \nonumber
\end{align}
The orientation reward is defined using the intrinsic $\SO$ trace error \cite{seo2023geometric}:
\begin{equation}
    r_{\mathrm{rot},t}
    =
    \exp\left(
    -\frac{
    \operatorname{mean}_{i\in\{l,r\}}
    \left(\operatorname{tr}(I_3 - R_{DE_i,t})\right)}
    {\sigma_R}
    \right).
\end{equation}

The residual policy is further regularized through penalties on the arm residual magnitude $\|a_{\mathrm{RL,arm},t}\|_2^2$, sampled-action rate $\|\Delta a_{t}\|_2^2$, and mean-action acceleration $\|\Delta^2\mu_t\|_2^2$, where $\mu_t=\mu_\theta(o_t)$ denotes the policy mean action and $\Delta^2\mu_t=\mu_t-2\mu_{t-1}+\mu_{t-2}$.



Locomotion is coordinated using the periodic gait state $\phi_t$, with the left and right leg phases offset by $\pi$. The gait phase also defines complete gait cycles $\mathcal{G}_k$, over which a signal $e_t$ is averaged as
\begin{equation}
    \bar e_k =
    {1}/{|\mathcal{G}_k|} \cdot
    \sum\nolimits_{t\in\mathcal{G}_k} e_t .
\end{equation}
Three drift penalties are introduced to suppress undesired planar motion. Under pure angular-velocity commands, \emph{turn-in-place drift} penalizes unintended planar translation. Under pure linear-velocity commands, \emph{heading drift} penalizes gait-cycle-averaged yaw rate, while \emph{lateral drift} penalizes gait-cycle-averaged velocity perpendicular to the commanded direction. Gait-cycle averaging avoids penalizing periodic locomotion oscillations that do not produce net drift.

\subsubsection{Curriculum}
\begin{table}[t]
    \setlength\doublerulesep{0.5pt}
    \renewcommand\tabularxcolumn[1]{m{#1}}
    \scriptsize
    \centering
    \setlength{\tabcolsep}{2.0pt}
    \renewcommand{\arraystretch}{1.05}
    \caption{
    Progressive training curriculum of ResGAC.
    Command ranges denote maximum absolute $(v_x,v_y,\omega_z)$ commands.
    For stages E and F, $\sigma_p$ denotes $(\sigma_2,\sigma_1)$
    of the combined $L_2^2+L_1$ kernel.
    }
    \label{table:curriculum}
    \vspace{-4pt}

    \begin{tabularx}{\columnwidth}{
        >{\raggedright\arraybackslash\hsize=0.95\hsize}X
        *{6}{>{\centering\arraybackslash\hsize=0.80\hsize}X}
    }
    \toprule[1pt]\midrule[0.3pt]
        & A & B & C & D & E & F \\
    \midrule\midrule

    Task
        & \shortstack[c]{Stand\\Fixed}
        & \shortstack[c]{Stand\\Traj.}
        & \shortstack[c]{Walk\\Fixed}
        & \shortstack[c]{Walk\\Traj.}
        & \shortstack[c]{Walk\\Traj.}
        & \shortstack[c]{Walk\\Traj.} \\

    Iterations
        & $\leq 10$k
        & $\leq 10$k
        & $\leq 20$k
        & $50$k
        & $60$k
        & $60$k \\

    Cmd. Range
        & $0$
        & $0$
        & $.3/.3/.5$
        & $.7/.7/.7$
        & $.7/.7/.7$
        & $.7/.7/.7$ \\
    \addlinespace[1pt]
    $\sigma_{p}$
        & $.02$
        & $.01$
        & $.02$
        & \shortstack[c]{$.005\!\rightarrow$\\$.001$}
        & $.001/.01$
        & $.001/.01$ \\
    \addlinespace[1pt]
    $\sigma_R$
        & $.20$
        & $.10$
        & $.10$
        & \shortstack[c]{$.10\!\rightarrow$\\$.02$}
        & $.02$
        & $.02$ \\
    \addlinespace[1pt]
    $\sigma_v$
        & $.25$
        & $.25$
        & $.25$
        & \shortstack[c]{$.25\!\rightarrow$\\$.10$}
        & $.10$
        & $.10$ \\

    Init. $q$ scale
    & $[.9,1.1]$
    & $[.85,1.15]$
    & $[.85,1.15]$
    & $[.7,1.3]$
    & $[.7,1.3]$
    & $[.7,1.3]$ \\

    $w_{\mathrm{rot}}$
        & $2.5$ & $2.5$ & $2.5$ & $2.5$ & $3.0$ & $3.0$ \\
    
    $w_{\mathrm{res}}$
        & $-.10$ & $-.10$ & $-.10$
        & \shortstack[c]{$-.10\!\rightarrow$\\$-.05$}
        & $-.05$ & $-.05$ \\
    
    $w_{\mathrm{pose}}$
        & $-.5$ & $-.5$ & $-.5$ & $-1.0$ & $-1.0$ & $-1.0$ \\
    
    $w_{\dot q,\mathrm{arm}}$
        & $-.0125$ & $-.0125$ & $-.0125$
        & $-.02$ & $-.05$ & $-.05$ \\
    
    $w_{\mathrm{twist}}$
        & $0$ & $0$ & $0$
    & $-.02$ & $-.05$ & $-.05$ \\

    \midrule[0.3pt]\bottomrule[1pt]
    \end{tabularx}
    \vspace{-4pt}
\end{table}
A progressive curriculum is employed as summarized in Table~\ref{table:curriculum}. Training begins with fixed-target standing, where the residual policy first learns to cooperate with the nominal GAC while maintaining balance. The subsequent stages progressively introduce trajectory variation and locomotion, tighten the tracking objective, and finally broaden the task-space command distribution to the ranges required for teleoperation and Mime-man. End-effector trajectories are generated directly in task space from feasible arm configurations, without retargeted whole-body human-motion data.

\subsubsection{Symmetry Augmentation and Regularization}
We exploit the left-right morphological symmetry of the humanoid through sagittal reflection \cite{Amazon_FAR_and_Abbeel_Holosoma}. Let $\mathcal{M}$ denote the full reflection operator, with Cartesian component $M=\operatorname{diag}(1,-1,1)$, which exchanges the left and right joints and applies the corresponding sign changes to directional quantities. 
Each replay transition is augmented with a mirrored counterpart obtained by applying $\mathcal{M}$ consistently to all reflection-dependent quantities.
To further encourage symmetric behavior, the actor is regularized to be equivariant and the critic invariant under $\mathcal{M}$ \cite{seo2025se}:
\begin{equation}
\begin{aligned}
\mathcal{L}^{\mathrm{actor}}_{\mathrm{sym}}
&=
\|\mu_\theta(\mathcal{M}\circ o_t)
-\mathcal{M}\circ\mu_\theta(o_t)\|_2^2,\\
\mathcal{L}^{\mathrm{critic}}_{\mathrm{sym}}
&=
\|Q_\theta(o_t,a_t)
-Q_\theta(\mathcal{M}\circ o_t,\mathcal{M}\circ a_t)\|_2^2,
\end{aligned}
\end{equation}
where $\circ$ denotes an action of $\mathcal{M}$ to the corresponding set.
These terms are added to the corresponding RL objectives.

\section{Experiments}
The proposed ResGAC is evaluated on the Unitree G1 humanoid through four \textbf{real-world experiments}: $\SE$ end-effector pose trajectory tracking, arm stabilization during walking, peg-in-hole task, and world-frame end-effector tracking.

\subsection{Implementation Details}
Training is conducted in \texttt{IsaacSim} using the Holosoma framework \cite{Amazon_FAR_and_Abbeel_Holosoma}, with FastSAC \cite{seo2025learning} used for residual policy optimization. A total of $4096$ parallel environments are trained on a single NVIDIA L40S GPU, and the complete six-stage curriculum requires approximately $16$ hours. Both GAC and the residual policy operate at $50\;\mathrm{Hz}$, and the resulting joint-position commands are tracked by the low-level PD controllers of the Unitree G1 with the PD control gains adopted from BeyondMimic \cite{liao2025beyondmimic}. At deployment, robot kinematics are evaluated using \texttt{pinocchio} \cite{carpentier-sii19}. For the relative velocity construction described in Sec.~\ref{sec:reference_frame}, the floating-base angular velocity is obtained directly from the G1 IMU, while the linear base-velocity components are omitted since they are not directly measured at deployment.

The GAC parameters are set to
$M=\operatorname{diag}(1.5I_3,2.25I_3)$,
$K_p=K_R=5000I_3$, and
$K_d$ is set to yield damping ratio of $2$.
The damped pseudo-inverse and null-space gains are
$\lambda=0.0025$ and $k_{\mathrm{null}}=50$,
respectively. The arm residual uses $s_r=0.75$ and $c_r=1.0$.

Additional low-level implementation details and experimental videos are available on the project website.

\subsection{Manipulator Tracking Performance}
\begin{table*}[t]
    \setlength\doublerulesep{0.5pt}
    \scriptsize
    \centering
    \setlength{\tabcolsep}{2.0pt}
    \renewcommand{\arraystretch}{1.08}
    \caption{End-effector tracking performance during standing in real-world hardware experiment. 
    Errors are reported as mean $\pm$ standard deviation of the per-sample error magnitudes, aggregated over temporal samples and both arms. For the setpoints tracking, only steady-state samples after each command transition are included. Lower values indicate better tracking performance, with the best and second-best results shown in bold and underlined, respectively. The command reference frame is the pelvis frame $\{P\}$ for all controllers except SONIC-Teleop-v1.1, which uses the heading frame $\{H\}$.}
    \label{table:tracking_benchmark}
    \vspace{-4pt}

    \begin{tabular*}{\textwidth}{
        @{\extracolsep{\fill}}
        l
        cc cc cc cc
        @{}
    }
    \toprule[1pt]\midrule[0.3pt]

    \multirow{2}{*}{Controller}
    & \multicolumn{2}{c}{Smooth}
    & \multicolumn{2}{c}{Setpoint}
    & \multicolumn{2}{c}{Infinity}
    & \multicolumn{2}{c}{Pick-and-Place} \\

    \cmidrule(lr){2-3}
    \cmidrule(lr){4-5}
    \cmidrule(lr){6-7}
    \cmidrule(lr){8-9}

    & Pos. [mm] & Rot. [$^\circ$]
    & Pos. [mm] & Rot. [$^\circ$]
    & Pos. [mm] & Rot. [$^\circ$]
    & Pos. [mm] & Rot. [$^\circ$] \\

    \midrule\midrule

    E2E-RL
    & 43.6 $\pm$ 31.1
    & \underline{9.36 $\pm$ 7.47}
    & 33.9 $\pm$ 20.9
    & \underline{4.87 $\pm$ 3.24}
    & 33.6 $\pm$ 14.9
    & 8.36 $\pm$ 4.42
    & 35.5 $\pm$ 31.6
    & \underline{6.02 $\pm$ 1.56} \\
    
    \addlinespace[1.5pt]

    GAC+Decoupled-RL
    & 98.5 $\pm$ 32.5
    & 14.80 $\pm$ 6.16
    & 94.5 $\pm$ 24.9
    & 13.31 $\pm$ 3.78
    & 50.6 $\pm$ 12.6
    & 9.47 $\pm$ 1.87
    & 74.1 $\pm$ 37.5
    & 12.25 $\pm$ 3.68 \\

    \addlinespace[1.5pt]

    SONIC-IK-WBT
    & 41.7 $\pm$ 27.6
    & 18.86 $\pm$ 13.37
    & 25.1 $\pm$ 17.6
    & 15.42 $\pm$ 11.01
    & 32.3 $\pm$ 32.4
    & 14.06 $\pm$ 13.01
    & 21.9 $\pm$ 5.9
    & 15.31 $\pm$ 5.94 \\

    \addlinespace[1.5pt]

    SONIC-Teleop
    & 39.6 $\pm$ 40.3
    & 23.23 $\pm$ 20.52
    & 25.3 $\pm$ 19.6
    & 18.32 $\pm$ 11.56
    & 18.4 $\pm$ 9.0
    & 11.00 $\pm$ 5.40
    & 17.3 $\pm$ 9.9
    & 24.63 $\pm$ 9.09 \\

    \addlinespace[1.5pt]

    SONIC-Teleop-v1.1
    & \underline{34.3 $\pm$ 29.0}
    & 17.60 $\pm$ 26.86
    & \underline{19.7 $\pm$ 9.6}
    & 8.65 $\pm$ 8.63
    & \underline{15.3 $\pm$ 6.4}
    & \underline{6.44 $\pm$ 2.77}
    & \underline{14.5 $\pm$ 3.9}
    & 6.23 $\pm$ 3.17 \\

    \addlinespace[1.5pt]

    \textbf{ResGAC (Ours)}
    & \textbf{22.4 $\pm$ 25.0}
    & \textbf{5.29 $\pm$ 7.00}
    & \textbf{6.5 $\pm$ 3.7}
    & \textbf{1.37 $\pm$ 0.76}
    & \textbf{14.1 $\pm$ 7.1}
    & \textbf{3.63 $\pm$ 2.15}
    & \textbf{6.6 $\pm$ 2.7}
    & \textbf{1.67 $\pm$ 0.75} \\

    \midrule[0.3pt]\bottomrule[1pt]
    \end{tabular*}
    \vspace{-4pt}
\end{table*}

We first evaluate $\SE$ end-effector pose tracking against five representative baselines:
\begin{itemize}[leftmargin=*]
    \item \textbf{E2E-RL:} An end-to-end RL policy trained with the same recipe as ResGAC, but without the nominal GAC controller.
    \item \textbf{GAC+Decoupled-RL:} The arms are controlled only by GAC, while the remaining whole-body joints are controlled by RL.
    \item \textbf{SONIC-IK-WBT:} Upper-body joint targets are obtained through inverse kinematics (IK) and tracked using the SONIC whole-body tracking model.
    \item \textbf{SONIC-Teleop:} The original three-point SONIC teleoperation policy using head and bilateral wrist targets.
    \item \textbf{SONIC-Teleop-v1.1:} The updated three-point SONIC teleoperation policy using a heading-frame representation.
\end{itemize}

All controllers are evaluated on four trajectory-tracking scenarios:
\begin{itemize}[leftmargin=*]
    \item \textbf{Smooth arm-sweep:} A fast and smooth trajectory spanning a broad arm workspace.
    \item \textbf{Setpoint arm-sweep:} A sequence of discrete full-arm-range setpoints used to evaluate step-response behavior. 
    \item \textbf{Infinity:} An infinity-shaped trajectory with simultaneous position and moderate orientation variation.
    \item \textbf{Pick-and-Place:} A slower task-space trajectory sampled from a teleoperated pick-and-place demonstration.
\end{itemize}

We first evaluate end-effector tracking in a standing configuration to isolate manipulator tracking precision from locomotion-induced disturbances.
All evaluation trajectories are held out from training; the training curriculum covers the corresponding task-space ranges and motion classes, but does not include any of the trajectories used for evaluation.
For a fair comparison, all baselines are evaluated in the pelvis frame when supported. SONIC-Teleop-v1.1 is the only exception, as its released policy uses a heading-frame representation: therefore, its errors are reported in their own reference frame. Position and orientation errors are reported as mean $\pm$ standard deviation over temporal samples and both arms. For the setpoint arm-sweep trajectory, only steady-state samples are reported. 

The tracking results are summarized in Table~\ref{table:tracking_benchmark}. 
ResGAC achieves the lowest position and orientation errors on all four trajectories, reaching sub-centimeter position error ($6.5\;\mathrm{mm}$ and $6.6\;\mathrm{mm}$) and below $2^\circ$ orientation error ($1.37^\circ$ and $1.67^\circ$) on the setpoint and pick-and-place trajectories, corresponding to $54–67\%$ lower position error and approximately $72\%$ lower orientation error than the best baseline.
The benefit of the structured residual formulation is particularly evident from the two architectural ablations. E2E-RL, in which the nominal geometric feedback is removed, exhibits consistently larger tracking errors, demonstrating the benefit of explicit task-space feedback. Conversely, GAC+Decoupled-RL, in which the arm residual is removed, performs substantially worse than ResGAC, indicating that geometric feedback alone is insufficient to compensate for effects such as gravity and whole-body dynamic coupling.

\begin{figure}[t!]
    \centering
    \includegraphics[width=0.9\linewidth]{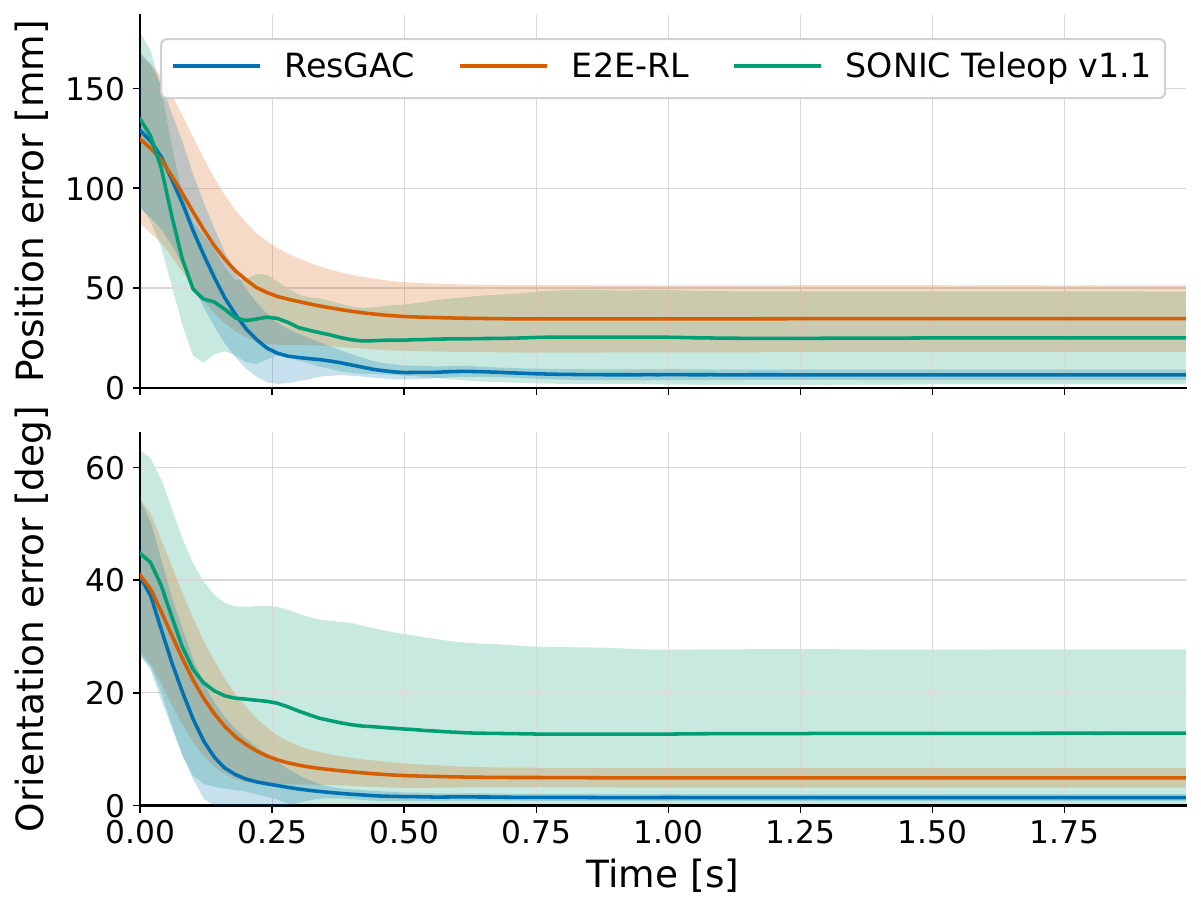}
    \vspace{-5pt}
    \caption{End-effector error convergence for the setpoint arm-sweeping trajectory. The $25$ setpoint intervals are aligned at each command transition ($t=0$) and aggregated. Solid curves indicate the mean position and orientation error magnitudes, while shaded regions indicate one standard deviation.}
    \label{fig:nonsmooth_eef_comparison}
\end{figure}

The setpoint arm-sweep trajectory further evaluates the transient response under discrete command changes. As shown in Fig.~\ref{fig:nonsmooth_eef_comparison}, ResGAC exhibits faster convergence together with lower steady-state position and orientation errors than the representative E2E-RL and SONIC-Teleop-v1.1 baselines.

\subsection{End-effector Stabilization via Frame Selection}

\begin{table}[t]
    \footnotesize
    \centering
    \setlength{\tabcolsep}{2.5pt}
    \renewcommand{\arraystretch}{1.05}
    \setlength\doublerulesep{0.3pt}
    \vspace{-6pt}
    \caption{
    Comparison of pelvis-frame $P$ and ground-attached heading-frame $H_0$ representations during walking arm holding using ResGAC.
    ``EEF MAE'' denotes the mean absolute error (MAE) from the fixed world-frame hold target, while ``Ratio'' denotes the motion-transmission ratio
    $\rho=\mathrm{EEF~Std.}/\mathrm{Pelvis~Std.}$.
    Lower values indicate better world-frame holding and weaker transmission, respectively.
    }
    \label{table:frame_selection}
    \vspace{-4pt}
    \begin{tabular*}{\columnwidth}{@{\extracolsep{\fill}} l c c c c c c @{}}
        \toprule
        \midrule
        & \multicolumn{3}{c}{Ground-Heading ($H_0$)} & \multicolumn{3}{c}{Pelvis ($P$)} \\
        \cmidrule(lr){2-4} \cmidrule(lr){5-7}
        Metric & Pelvis Std. & EEF MAE & Ratio & Pelvis Std. & EEF MAE & Ratio \\
        \midrule
        \midrule
        Roll [$^\circ$]
        & 1.218 & \textbf{0.807} & \textbf{0.790}
        & 1.322 & 1.113 & 0.958 \\

        Pitch [$^\circ$]
        & 3.794 & 2.600 & 0.954
        & 4.107 & 2.757 & 0.964 \\

        $z$ [mm]
        & 6.280 & \textbf{12.56} & \textbf{0.884}
        & 6.219 & 22.50 & 1.068 \\
        \midrule
        \bottomrule
    \end{tabular*}
    \vspace{-4pt}
\end{table}

The effect of manipulation-frame selection is evaluated through a static arm-holding task during forward and backward walking. Two ResGAC policies are compared using the pelvis frame $P$ and the ground-attached heading frame $H_0$ as their manipulation reference frames. In both cases, a fixed end-effector pose is commanded in the corresponding reference frame.
For the static-arm-holding and world-frame tracking experiments, a global wrapper is additionally used to generate locomotion commands from the estimated planar pelvis pose $(x,y,\psi)\in SE(2)$. 
The pelvis pose is estimated using OptiTrack, with marker-to-pelvis-frame calibration, and a PD-like feedback law converts the planar pose error into linear and angular velocity commands. The wrapper determines only the locomotion command, while end-effector tracking and whole-body coordination remain handled by ResGAC.



To quantify world-frame end-effector stabilization during walking, we report both the end-effector mean absolute error (MAE) from the fixed hold target and the transmission of pelvis motion to the end effector. For each component $x\in\{\mathrm{roll},\mathrm{pitch},z\}$, the transmission ratio is defined as 
$\rho_x=\sigma_{\mathrm{EEF},x}/\sigma_{P,x}$,
where $\sigma_{\mathrm{EEF},x}$ and $\sigma_{P,x}$ denote the standard deviations of the end-effector and pelvis motion, respectively. A lower $\rho_x$ indicates weaker transmission of pelvis motion.


The results are summarized in Table~\ref{table:frame_selection}. The pelvis-motion statistics remain comparable between the two policies, indicating similar locomotion-induced disturbance levels. For roll, the $H_0$ representation reduces the transmission ratio from $0.958$ to $0.790$ and the EEF MAE from $1.11^\circ$ to $0.81^\circ$, demonstrating reduced reference-induced roll transmission. Pitch is largely unaffected by frame selection, with similar transmission ratios of $0.954$ and $0.964$, because the lever-like arm configuration primarily transmits locomotion-induced impacts in pitch. For $z$, $H_0$ reduces the EEF MAE from $22.5$ mm to $12.6$ mm and the transmission ratio from $1.068$ to $0.884$. Overall, the results demonstrate improved attenuation of locomotion-induced end-effector motion when ResGAC uses $H_0$ as the manipulation reference frame.

\subsection{Peg-in-Hole Task}
We further evaluate whether the end-effector tracking accuracy is
sufficient for task-level manipulation through a standing peg-in-hole experiment, which isolates manipulator tracking accuracy from lower-body placement errors. The hole location is estimated from OptiTrack markers and used to construct the insertion target pose for both controllers. Since the marker-to-hole transform is not precisely calibrated, the commanded target contains a registration error with respect to the true hole pose. To accommodate this uncertainty, the task uses a $10$-mm diametral clearance between the peg ($25\;\mathrm{mm}$ diameter) and hole ($35\;\mathrm{mm}$ diameter). Each controller is evaluated over $20$ trials.

ResGAC succeeds in $\mathbf{18/20}$ trials ($\mathbf{90\%}$), compared with $\mathbf{10/20}$ ($\mathbf{50\%}$) for SONIC-Teleop-v1.1. The two ResGAC failures occur when the registration error is compounded by temporary OptiTrack marker occlusion.
For SONIC, insufficient tracking accuracy is the primary failure mode, 
with the same registration and occlusion uncertainties also present.
These results show that the improved end-effector tracking accuracy of ResGAC enables a higher task-level success rate under the same externally estimated insertion target.


\subsection{World-frame Tracking: Mime-Man Demo}
We finally evaluate world-frame end-effector tracking through the Mime-man demo shown in Fig.~\ref{fig:ResGAC_main}(c). The end effectors are first moved to designated world-frame poses and are subsequently commanded to remain fixed in the world while the humanoid translates and rotates its body. The global wrapper described in Sec.~IV-C generates the locomotion commands from the OptiTrack-based pelvis localization.

ResGAC operates with end-effector commands represented in the ground-attached heading frame $H_0$. Therefore, a desired world-frame pose $g_{WD}$ needs to be converted at each control step as $g_{H_0D}
    =
    g_{WH_0}^{-1}\!\cdot g_{WD}.$
Both ResGAC and E2E-RL are evaluated using the same $H_0$ reference frame formulation, global wrapper, and world-to-$H_0$ command conversion. As the humanoid moves, this transformation produces a time-varying command in $H_0$. To reduce the resulting closed-loop tracking lag, the end-effector tracking dynamics are approximated by a second-order system $G(s) = \tfrac{\omega_n^2}{s^2+2\zeta\omega_n s + \omega_n^2}$, and a causal inverse-dynamics compensator $G^{-1}(s)$, realized in the time-domain, is applied to the converted reference. 
The required first- and second-order reference derivatives are computed numerically for causality.

\begin{figure}[t!]
    \centering
    \includegraphics[width=1.0\linewidth]{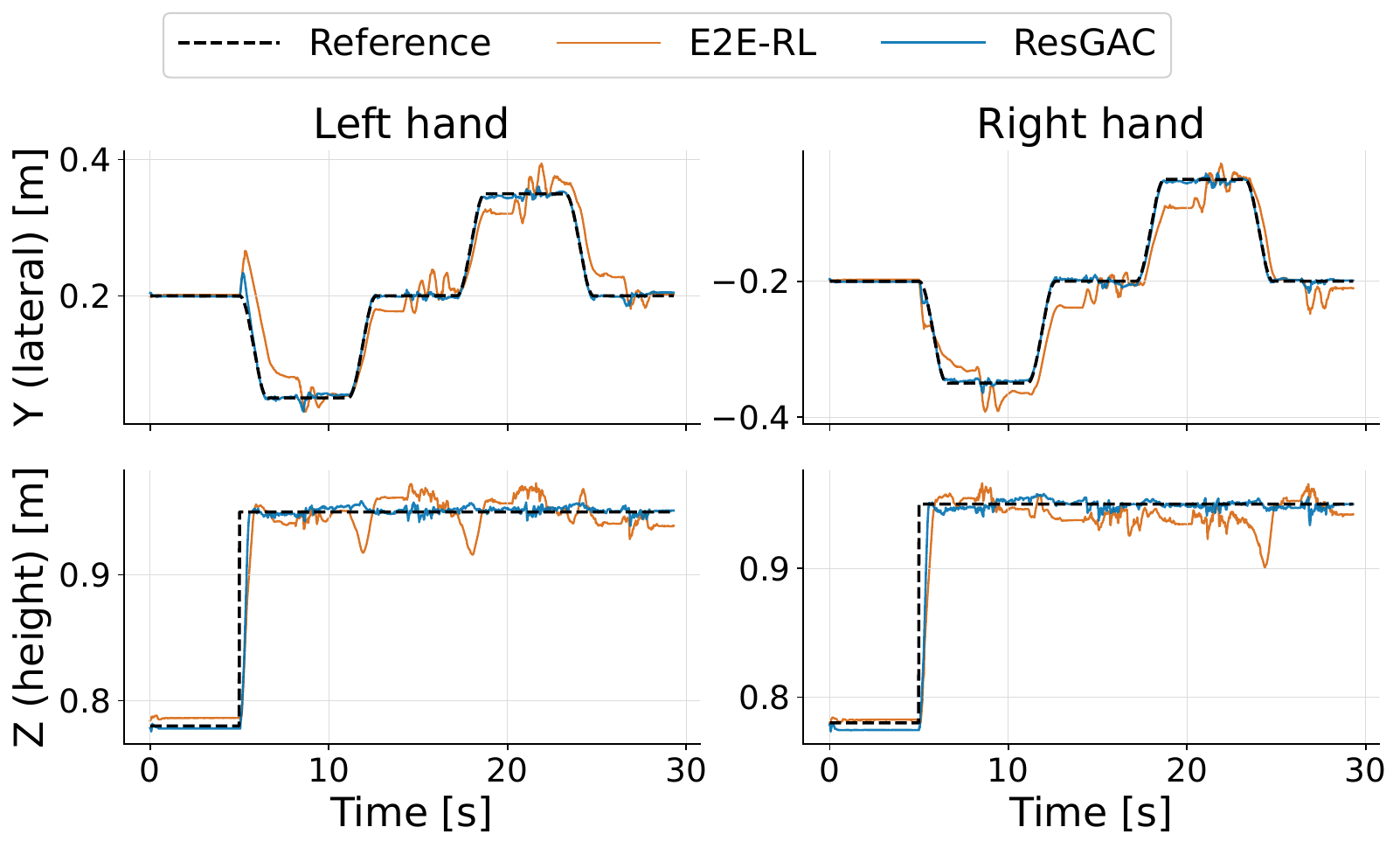}
    \vspace{-18pt}
    \caption{World-frame $YZ$ tracking in the Mime-man demo, comparing ResGAC and E2E-RL. 
    ResGAC achieves position and orientation MAEs of $\mathbf{8.8\;\mathrm{mm}}$, $\mathbf{2.69^\circ}$, respectively, compared with $\mathbf{30.0\;\mathrm{mm}}$ and $\mathbf{13.75^\circ}$ for E2E-RL.
    %
    }
    \label{fig:mime_comparison}
\end{figure}

The overlaid sequence in Fig.~\ref{fig:ResGAC_main}(c) qualitatively shows that the end effectors remain approximately fixed in the world while the lower body moves. The corresponding world-frame $YZ$ trajectories are shown in Fig.~\ref{fig:mime_comparison}. 
ResGAC maintains substantially closer tracking of the commanded world-frame pose, achieving position and orientation MAEs of $\mathbf{8.8\;\mathrm{mm}}$ and $\mathbf{2.69^\circ}$, respectively, compared with $\mathbf{30.0\;\mathrm{mm}}$ and $\mathbf{13.75^\circ}$ for E2E-RL.
Even with the world-frame reference compensator tuned for E2E-RL, the E2E-RL exhibits substantial tracking degradation, demonstrating the benefit of the structured geometric feedback in maintaining accurate world-frame end-effector pose tracking during body motion.

\section{Conclusion}
We presented ResGAC, a whole-body humanoid control framework for high-precision end-effector pose tracking. By combining geometric admittance control (GAC) with residual RL in a common joint-position action space, ResGAC retains structured geometric task-space feedback while learning the corrections required for locomotion, balance, and unmodeled whole-body dynamics. 
The left-invariant geometric formulation further allows the same GAC law to be used consistently across manipulation reference frames, while the proposed ground-attached heading frame prevents pelvis roll, pitch, and vertical motion from propagating to the desired end-effector pose via reference transformation.
Real-world experiments on the Unitree G1 demonstrate improved translational and rotational tracking over representative baselines, end-effector stabilization during locomotion through frame selection, higher task-level success in standing peg-in-hole manipulation, and accurate world-frame pose tracking during body motion. 
These results demonstrate the applicability of ResGAC as a precise $\SE$ task-space interface in the whole-body humanoid control including loco-manipulation.

\bibliographystyle{IEEEtran}
\bibliography{references}

\end{document}